%% file: main.tex
\documentclass[10pt,twocolumn,letterpaper]{article}

\usepackage[pagenumbers]{cvpr} 
\usepackage{multirow} 
\usepackage{caption}
\usepackage{makecell}
\usepackage{titletoc}
\usepackage[misc]{ifsym}

\input{preamble}
\definecolor{cvprblue}{rgb}{0.21,0.49,0.74}
\usepackage[pagebackref,breaklinks,colorlinks,allcolors=cvprblue]{hyperref}

\def\paperID{} 
\def\confName{CVPR}
\def\confYear{2026}

\title{Remote Sensing Image Super-Resolution for Imbalanced Textures: A Texture-Aware Diffusion Framework}

\author{Enzhuo Zhang \quad Sijie Zhao \quad Dilxat Muhtar \quad Zhenshi Li \quad Xueliang Zhang\thanks{Corresponding author.} \quad Pengfeng Xiao\\
Nanjing University, China\\
{\tt\small zenzhuo@smail.nju.edu.cn, zxl@nju.edu.cn}
}

\begin{document}

\maketitle
\input{sec/0_abstract}    
\input{sec/1_intro}
\input{sec/2_relat}

\input{sec/3_metho}
\input{sec/4_exp}
\input{sec/5_conclu}

{
    \small
    \bibliographystyle{ieeenat_fullname}
    \bibliography{main-main}
}

\end{document}

%% file: sec/0_abstract.tex
\begin{abstract}

Generative diffusion priors have recently achieved state-of-the-art performance in natural image super-resolution, demonstrating a powerful capability to synthesize photorealistic details. However, their direct application to remote sensing image super-resolution (RSISR) reveals significant shortcomings. Unlike natural images, remote sensing images exhibit a unique texture distribution where ground objects are globally stochastic yet locally clustered, leading to highly imbalanced textures. This imbalance severely hinders the model's spatial perception. To address this, we propose TexADiff, a novel framework that begins by estimating a Relative Texture Density Map (RTDM) to represent the texture distribution. TexADiff then leverages this RTDM in three synergistic ways: as an explicit spatial conditioning to guide the diffusion process, as a loss modulation term to prioritize texture-rich regions, and as a dynamic adapter for the sampling schedule. These modifications are designed to endow the model with explicit texture-aware capabilities. Experiments demonstrate that TexADiff achieves superior or competitive quantitative metrics. Furthermore, qualitative results show that our model generates faithful high-frequency details while effectively suppressing texture hallucinations. This improved reconstruction quality also results in significant gains in downstream task performance. The source code of our method can be found at https://github.com/ZezFuture/TexAdiff.

\end{abstract}

%% file: sec/1_intro.tex
\section{Introduction}
\label{sec:intro}

\begin{figure}[t]
\centering
\includegraphics[width=1 \columnwidth]{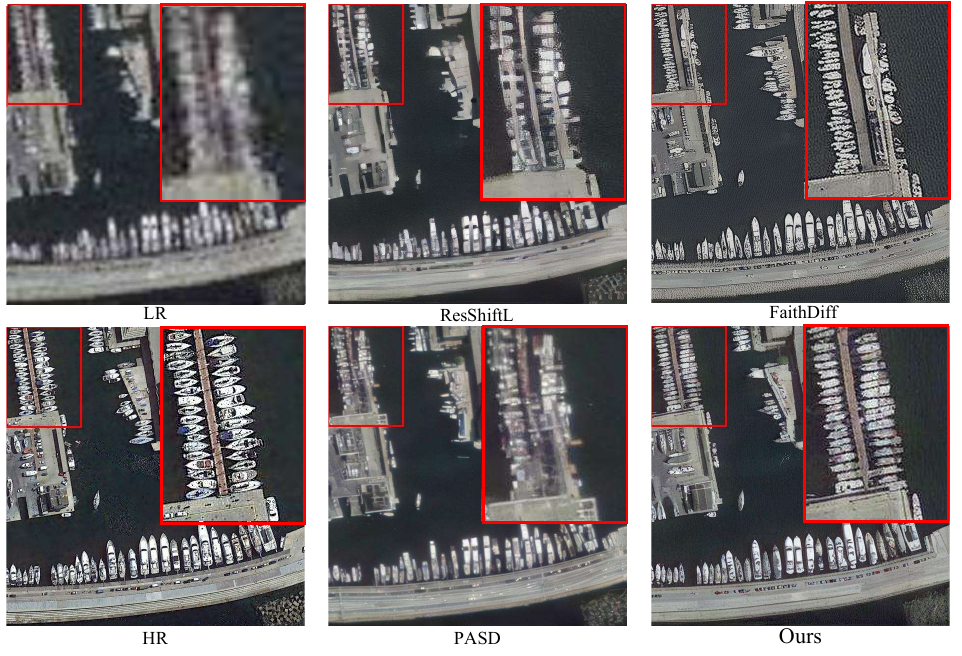} 
\caption{Our method produces faithful, fine-grained details in the texture-rich ship regions, while avoiding stripe artifacts in the texture-sparse water regions.}
\label{fig1}
\end{figure}

Remote sensing image super-resolution (RSISR) is a fundamental task in remote sensing image processing, aiming to reconstruct high-resolution (HR) images from low-resolution (LR) inputs. By enhancing spatial resolution, RSISR directly benefits critical downstream applications such as object detection~\cite{zhang2023superyolo}, semantic segmentation~\cite{chen2023large}, and change detection~\cite{liu2021super}. Despite the development of advanced network architectures and degradation modeling techniques, RSISR performance still suffers from severe degradation and poor generalization when confronted with unknown degradation types and imbalanced texture distributions inherent in real-world remote sensing images.

Recently, diffusion models, particularly those leveraging powerful pre-trained text-to-image (T2I) diffusion priors~\cite{chen2024pixart,podell2024sdxl}, have demonstrated remarkable success in natural image super-resolution. By incorporating diverse conditioning signals and employing carefully designed architectures, these models can generate sharp and realistic details. While this paradigm offers a promising direction for tackling unknown degradations in RSISR, it fails to adequately address the unique challenge of imbalanced texture distributions inherent to remote sensing images.

Remote sensing images exhibit pronounced spatial heterogeneity: a subset of texture-rich regions (e.g., road networks, built-up areas) carries most of the high-frequency energy, whereas extensive texture-sparse expanses (e.g., water bodies, snow-covered fields) are typically less structurally complex. The texture distribution is highly imbalanced and locally clustered, forming large and contiguous regions of high- and low-frequency content. From a global perspective, the positions of these clusters are scene-dependent and lack a global positional prior; the same type of structure may appear in different absolute positions across images. 
However, existing diffusion-based RSISR methods typically apply a uniform restoration intensity across the entire image, ignoring this intrinsic spatial heterogeneity. This spatially-invariant approach leads to two predictable failure modes. First, in texture-sparse regions, the model applies excessive restoration, inventing redundant or hallucinated details that introduce artifacts rather than improving quality. Second, in texture-rich regions, the restoration effort is insufficient, resulting in blurry reconstructions with missing details. As shown in Figure~\ref{fig1}, state-of-the-art methods like FaithDiff~\cite{chen2025faithdiff} and PASD~\cite{yang2024pixel} suffer from these issues; their uniform processing generates unrealistic textures in simple areas while failing to recover fine details in complex ones. These methods lack the mechanisms to recognize regional texture disparities in the LR images and adaptively allocate their representational capacity.

To address this challenge, we propose Texture-aware Diffusion (TexADiff), a novel super-resolution framework designed to handle the texture heterogeneity of remote sensing images adaptively. TexADiff begins by constructing a Relative Texture Density Map (RTDM) that quantifies the underlying pixel-wise texture distribution of the target SR image. This map effectively highlights both texture-rich and texture-sparse regions, enabling differentiated restoration. We leverage the RTDM in three synergistic ways: i) as a spatial condition to explicitly inform the diffusion model of regional texture disparities; ii) as a loss modulator in the training objective, compelling the model to allocate greater representational capacity to texture-rich regions; and iii) as a schedule adapter for the denoising process, assigning region-specific step counts to control the amount of detail synthesized. This multi-faceted strategy focuses on where detail is needed, producing finer and more faithful details while preventing artifacts in simpler regions.


Our framework introduces the RTDM as an additional condition alongside the LR image, posing a significant challenge in the efficient fusion of multiple heterogeneous inputs. A common strategy is to employ a separate ControlNet~\cite{zhang2023adding} for each condition, but this approach inflates model parameters, increases inference latency, incurs a substantially larger training-time GPU memory footprint. Inspired by ControlNext~\cite{peng2024controlnext}, we introduce a lightweight MiniControlNet which encodes and fuses all conditional inputs within a single, efficient branch. Furthermore, we selectively unfreeze a portion of the base diffusion model's parameters during training, which helps bridge the domain gap between its natural image priors and the specific characteristics of remote sensing data.

To summarize, our contributions are threefold:

\begin{itemize}

\item We propose TexADiff to address the spatial heterogeneity in RSISR, which is a texture-aware framework that adaptively handles the imbalanced and spatially clustered texture distributions in remote sensing images.

\item We introduce a lightweight MiniControlNet to efficiently fuse heterogeneous spatial conditions and propose a set of synergistic adaptive strategies that leverage a predicted texture map to guide the model's focus, loss, and denoising schedule.

\item We demonstrate through extensive experiments that TexADiff achieves better perceptual metrics on most benchmark remote sensing super-resolution datasets, producing perceptually superior results with more faithful and realistic details.

\end{itemize}

%% file: sec/2_relat.tex
\section{Related Work}

\subsection{Image Super-Resolution}

In SR, two key aspects are degradation modeling and network architecture design. From the perspective of degradation modeling, early methods~\cite{dong2014learning, lim2017enhanced} simulate LR images by applying simple bicubic downsampling to HR images, which leads to a large domain gap for real-world images. To construct more realistic synthetic datasets, subsequent approaches have achieved promising results either by estimating degradation kernels from real LR images ~\cite{bell2019blind}, or by using complex handcrafted degradation pipelines ~\cite{wang2021real, zhang2021designing}. From the perspective of network architecture design, PSNR-oriented super-resolution methods~\cite{shi2016real,liang2021swinir} often produce over-smoothed results due to their pixel-level loss. GAN-based approaches~\cite{ledig2017photo, wang2018esrgan}, while improving perceptual quality, are often prone to mode collapse and artifacts. 
Diffusion models, known for their ability to generate high-frequency details, have been introduced into super-resolution tasks. SRDiff~\cite{li2022srdiff} is the first to apply diffusion probabilistic models to this task. ResShiftL~\cite{yue2024efficient} further improves efficiency by leveraging a residual-shifting Markov chain and a tailored noise schedule. However, these diffusion-based methods generally lack the incorporation of generative priors, which limits their performance.

\subsection{Pre-trained T2I Diffusion Priors for RSISR}

Owing to the strong generative priors of pre-trained T2I diffusion models, recent super-resolution methods built upon them have shown promising performance in natural image super-resolution tasks. Recent research has focused on improving performance through advancements in model architecture~\cite{wang2024exploiting, yang2024pixel}, data scale~\cite{yu2024scaling, ai2024dreamclear}, control conditions~\cite{sun2024coser,wang2025semantic},  fine-tuning methods~\cite{chen2025faithdiff} and inference efficiency~\cite{wang2024sinsr, yue2025arbitrary}.
Different from these efforts, we endow the model with texture awareness to generate high-resolution remote sensing images that are both high visual quality and physically faithful.

%% file: sec/3_metho.tex

\section{Methodology}

\begin{figure}[t]
\centering
\includegraphics[width=1 \columnwidth]{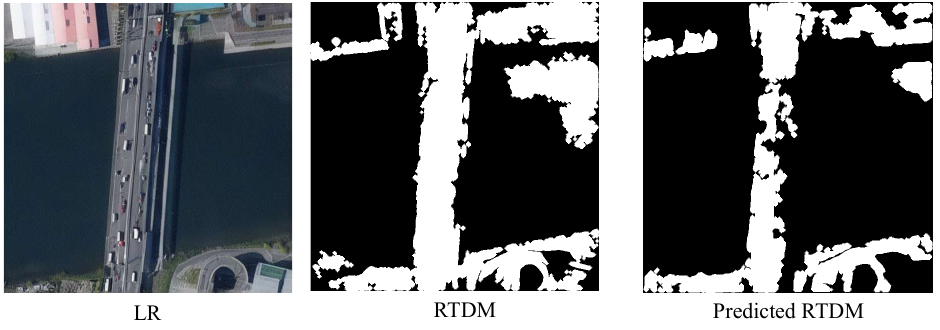} 
\caption{RTDM derived directly from the LR-HR pair during training and the model predicted RTDM at inference.}
\label{fig3}
\end{figure}

\begin{figure*}[t]
\centering
\includegraphics[width=2.1\columnwidth]{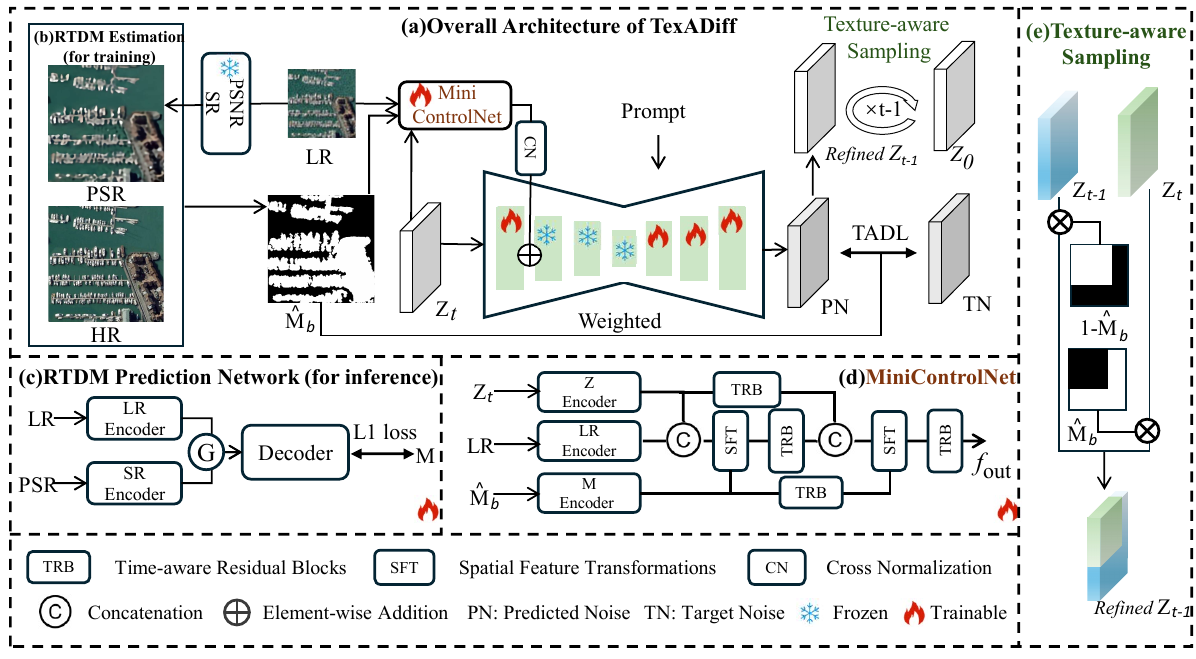} 
\caption{Architecture of proposed TexADiff. During training, the extracted RTDM is combined with the LR input and noisy latent via a MiniControlNet to form conditioning signals, while also modulating the training objective in a texture-aware manner. At inference time, since the HR image is unavailable, the RTDM is predicted by an RTDM prediction network.}
\label{fig_method}
\end{figure*}

\textbf{Framework Overview.} To enable texture-aware generation within the diffusion framework, we propose TexADiff, which consists of three key components:
i) an RTDM estimation and prediction module, which enables the identification of underlying texture distribution within the images, laying the foundation for subsequent processing; ii) a lightweight MiniControlNet capable of effectively integrating multiple conditional inputs, addressing the multi-condition guidance needs proposed in this study; and iii) a texture-aware diffusion denoising model, which leverages the RTDM to enable texture-aware denoising guided by texture distribution.
The overall architecture and detailed components are illustrated in Figure~\ref{fig_method}.

\subsection{RTDM Estimation and Prediction Module}

To provide a texture-aware prior for the denoising diffusion process, we introduce the Relative Texture Density Map (RTDM). The RTDM captures local texture density discrepancies between the HR $I^{HR}$ and LR $I^{LR}$, which correspond to the high-frequency details lost during the image degradation process, thereby characterizing the underlying pixel-wise texture distribution of the target SR image. The RTDM construction pipeline employs the direct estimation method during training and the prediction network during inference. 

Given that the LR input is both noise-contaminated and resolution-misaligned with the HR, we first employ a PSNR-oriented super-resolution model to produce a preliminary SR image $I^{PSR}$. PSNR-oriented models are particularly suited for this purpose, as they effectively suppress noise and tend to produce over-smoothed results in potentially texture-rich regions.
Local texture discrepancies can lead to differences in local statistics and perceptual divergence. This observation motivates the extraction of the RTDM. We adopt two complementary criteria: i) an SSIM-based contrast consistency term (CCT) utilizing local statistics, which is highly sensitive to degraded local textures, and ii) a spatial LPIPS response~\cite{zhang2018unreasonable}, which captures perceptual divergence while remaining robust to irregular or anisotropic textures,

\begin{gather}
M_{CCT}(i,j) = \text{CCT}(I^{PSR}, I^{HR}), \\
M_{SL}(i,j) = \text{Spatial LPIPS}(I^{PSR}, I^{HR}), 
\end{gather}
where $M_{CCT}(i,j)\in[0,1]$ denotes the local contrast consistency between the two images at location$(i,j)$, with lower values indicating stronger local texture discrepancies. $M_{SL}(i,j)\in[0,1]$ measures the perceptual dissimilarity at (i,j), where higher values correspond to greater perceptual divergence. 
Then, we invert $M_{SL}$ and combine them as:
\begin{equation}
M(i,j) = (1 - M_{SL}(i,j)) \cdot M_{CCT}(i,j),
\end{equation}
where lower values of ${M(i,j)}$ reflect stronger discrepancies in texture density. All maps share the $I^{HR}$ spatial resolution.

We empirically observed that guiding the model with the continuous-valued $M$ leads to unstable or chaotic outputs. To convert this into a clear and stable guidance signal, we apply threshold-based binarization, 
\begin{equation}
M_b(i, j) = 
\begin{cases}
1, & \text{if } M(i, j) \leq \tau \\
0, & \text{otherwise}
\end{cases}
\end{equation}
where $\tau$ is a threshold randomly sampled from the interval [0.35, 0.4] for each training batch. The impact of this hyperparameter is discussed in our ablation study. Next, we post‑process $M_b$ and downsample it to align with the latent-space resolution, obtaining the final binarized RTDM

\begin{equation}
\hat M_b = \mathcal{D}\!\big(\mathcal{P}(M_b)\big),
\end{equation}
where $\mathcal{P}$ denotes a post‑processing operator consisting of morphological erosion, dilation, and small‑component filtering. $\mathcal{D}$ is conducted using $8\times$ max-pooling.

During training, we directly utilize the $\hat{M}_b$ returned by the aforementioned procedure. At inference, however, $I^{HR}$ is unavailable, rendering this estimation method inapplicable.
To address this, as illustrated in Figure~\ref{fig_method}(c), we introduce an RTDM prediction network. It takes the LR and PSR images as inputs and encodes them through separate branches. A gating mechanism then allows features to selectively flow into a U-Net-style decoder, finally outputting a predicted RTDM.  Continuous-valued $M$(Eq. (1)–(3)) is used as a supervision pseudo-label to train the RTDM prediction network. The prediction loss is the mean absolute error ($L_1$ loss). The network outputs a continuous-valued map, which is subsequently binarized using the same method employed during training. Examples of RTDM obtained by the HR-based estimation method and the prediction network are shown in Figure~\ref{fig3}. We treat the RTDM generated by the estimation method as ground truth to evaluate the accuracy of the prediction method, as shown in Table~\ref{acc}. 

\begin{table}[htbp]
  \centering
  \small
  \caption{Accuracy of RTDM predictions, measured against the HR-based estimation.}
  \setlength{\tabcolsep}{2pt} 
  \begin{tabular}{ccccc}
    \toprule
    Datasets & RSC11~\cite{zhao2016feature} & LoveDA~\cite{wang2021loveda} & AID~\cite{xia2017aid} & DOTA~\cite{xia2018dota} \\
    \midrule
    Accuracy(\%)   & 78.47   & 79.56   & 71.01  & 71.97   \\
    \bottomrule
  \end{tabular}
\label{acc}
\end{table}

\subsection{MiniControlNet}

Previous works~\cite{zhang2023adding} often require multiple ControlNets to process different conditioning inputs. This increases inference cost and makes it harder to integrate multiple conditions effectively. To solve this, we adapt the lightweight ControlNext~\cite{peng2024controlnext} architecture and propose MiniControlNet, a more efficient alternative to the Multi-ControlNet approach that enables better integration of diverse conditional inputs. The design of MiniControlNet is inspired by ControlNext. As shown in Figure~\ref{fig_method}(d), its architecture includes parallel condition embedding branches, time-aware residual blocks, and Spatial Feature Transformation (SFT) layers~\cite{wang2018recovering}. The time-aware residual blocks accept time steps as input, while the parallel condition embedding encoders handle different conditioning inputs. The SFT layers inject the RTDM features into other feature streams. Following the approach in ControlNext, we use Cross Normalization to align the fused control features with the main branch features and inject the controls after the first U-Net block.

MiniControlNet reduces the control branch to just 20 million parameters, which is almost negligible compared to the large denoising U-Net. However, this compression limits its ability to guide the generation process effectively. To compensate for this limited capacity and prevent a fully frozen U-Net from inadequately adapting natural image priors to remote sensing images, as shown in Figure~\ref{fig_method}(a), we selectively unfreeze a subset of U-Net parameters, which include the first downsampling block and all upsampling blocks of the U-Net. We analyze the choice of unfrozen parameters in the ablation study.

\subsection{Texture-aware Denoising Diffusion Model}

We adopt multiple strategies to help the model perceive imbalanced and spatially clustered texture distributions and perform adaptive processing. 

Firstly, to explicitly guide the model to perceive the texture distributions of the images, the binarized RTDM $\hat{M}_b$ together with the LR image $I^{LR}$ and the current noisy latent $\hat{Z}_t$ are provided as conditional inputs to MiniControlNet. The resulting features are then added element-wise to the features of the denoising model.

We further refine the training objective of the denoising diffusion process. Empirically, we find that spatial structures apparent in pixel space are largely retained after VAE~\cite{kingma2014auto} encoding, i.e., in latent space. Recent diffusion‑based inpainting methods~\cite{lugmayr2022repaint} exploit this property by applying pixel‑space masks to steer local latent generation, confirming that spatial modulation in the latent space is both feasible and effective. Building on this insight, we propose the Texture-Aware Diffusion Loss (TADL), which strengthens the model’s noise prediction capability in texture-rich regions. Specifically, we assign higher weights to noise prediction errors in regions with high texture complexity, thereby reinforcing the model’s learning signal in those areas. The loss is formally defined as,


\begin{equation}\label{euq_tadl}
\begin{split}
\mathcal{L}_{TADL} &= \mathbb{E}_{z_0, c, t, \epsilon, I^{LR},\hat{M}_b} \Bigl[ \\
&\quad (1 + \alpha \hat{M}_b) \odot \bigl( \epsilon - \epsilon_\theta(z_t, c, t, I^{LR}, \hat{M}_b) \bigr)^2 \Bigr],
\end{split}
\end{equation}
where $\epsilon_\theta$ denotes the U-Net noise prediction network, $c$ is text prompt, $t$ is a diffusion timestep randomly sampled, $\alpha$ is an adjustable weight (set to 1), $\epsilon \sim N (0,1)$.

During the sampling process, early denoising steps form the global layout, whereas later iterations inject high-frequency details. To exploit this property, we introduce a texture‑aware dynamic sampling schedule. Since texture-sparse regions require fewer fine details, $\hat{M}_b$ is used to skip part of the late denoising steps for those regions. Specifically, during a predefined interval of late diffusion timesteps (e.g., $t \in [100, 500]$), we employ an alternating update strategy for the latent variables. As illustrated in Figure~\ref{fig_method}(e), the latents in texture-sparse regions, which are identified by the mask $\hat{M}_b$, are updated only every other step. In the intervening steps, these specific latents remain unchanged from their previous state, effectively pausing their refinement. This approach alternates between a spatially-selective update and a global update across the noisy latent.

%% file: sec/4_exp.tex
\begin{table*}[ht]{%
\centering
\caption{Quantitative comparison on the AID, DOTA-Test, LoveDA-Val and RSC11 datasets. Best results are in \textbf{bold}, second-best are \underline{underlined}. $\uparrow$ indicates that higher is better. $\downarrow$ indicates lower is better.}
\label{tab:sy}
\begin{tabular}{l|l|ccccccccccc}
\toprule
Datasets & Methods & PSNR$\uparrow$ & SSIM$\uparrow$ & LPIPS$\downarrow$ & DISTS$\downarrow$ &  NIQE$\downarrow$ & BRISQUE$\downarrow$ &  CLIP-IQA+$\uparrow$ \\
\midrule


\multirow{6}{*}{AID} 
& Real-ESRGAN    & \textbf{23.22} & \textbf{0.5280} &  0.4270 & 0.2409  & \textbf{4.19} & 24.51 & 0.5046 \\
& ResShiftL      & 23.03 & 0.4566 & 0.4504 & 0.2604  & 7.34& 26.18 &  0.5551 \\
& PASD         & \underline{23.20} & \underline{0.5061} & 0.4213 & 0.2066 &   \underline{4.97} & 25.79 &   0.5994   \\
& FaithDiff   & 22.79 & 0.4696 & \textbf{0.3752} & 0.2065  & 5.43 & 22.93 & 0.6134 \\
& Ours(thr=0.35)           &22.87   & 0.4696 & \underline{0.3788} & \textbf{ 0.1939} &  5.08 & \underline{ 21.96}  &  \underline{0.6149} \\
& Ours(thr=0.40)    & 22.62 & 0.4554 & 0.3823 & \underline{0.1953} &  5.00 & \textbf{ 20.63}  &  \textbf{0.6354} \\
\bottomrule

\multirow{6}{*}{LoveDA-Val} 
& Real-ESRGAN    & \textbf{26.88} & \textbf{0.6513}  & 0.4031 & 0.2244  & 5.46 & 42.36  & 0.3739 \\
& ResShiftL    & 25.48 & 0.5127 & 0.4992 & 0.2486  & 7.13 & \textbf{24.44}  & \textbf{0.5240}    \\
&  PASD  & 26.03 & \underline{0.5818}  & 0.3548 &  \textbf{0.1594}  & \textbf{5.26} & 36.00  & \underline{0.4718} &  \\
& FaithDiff   & \underline{26.33}    & 0.5791 & 0.3573 & 0.1873 &   5.96  &36.84 &   0.4008 \\
& Ours(thr=0.35)    &    26.34 &0.5779 &  \underline{0.3264} & 0.1766& 5.40 & 31.64 & 0.3919 \\
& Ours(thr=0.40)    &   26.15   & 0.5662 &  \textbf{0.3253} &  \underline{0.1751}& \underline{5.36} & \underline{31.02} &    0.4105   \\

\bottomrule

\multirow{6}{*}{DOTA-Test} 
& Real-ESRGAN  & \textbf{25.48} & \textbf{0.6298} & 0.3801 &  0.2358 &   \textbf{5.24} & \textbf{24.70}  & 0.4759
\\
& ResShiftL   & 24.99 & 0.5552 &  0.4515 &  0.2822 & 8.34 & \underline{26.19}  & 0.4808  \\
& PASD & \underline{25.08}   & \underline{0.5955} & 0.4155 &  0.2311  & \underline{5.98} & 26.62&  0.5476  \\
& FaithDiff & 24.62 & 0.5608 &  0.3448 & 0.2130 & 6.45  & 28.17 &  \textbf{0.5815}\\       
& Ours(thr=0.35)   &  24.74& 0.5711 & \textbf{0.3358}  & \textbf{0.1934} &  6.24 & 27.59 &  0.5677\\
& Ours(thr=0.40)   &  24.52&  0.5580 & \underline{0.3417}  & \underline{ 0.1978} &  6.28 &  27.51&\underline{0.5801}\\

\bottomrule

\multirow{6}{*}{RSC11} 
& Real-ESRGAN & \underline{21.57} & \textbf{0.4167} & 0.5528 &  0.2825 &   \textbf{4.33} & 24.10  &  0.4575 \\
& ResShiftL   & \underline{21.57} & 0.3723 &  0.5268 &  0.3014 & 7.83 & 27.43  & 0.4564  \\
& PASD &  \textbf{21.62}   & \underline{0.4086} & 0.5001 &  0.2746  & 6.21 & 39.42&  0.4773  \\
& FaithDiff & 20.65 & 0.3516 &  0.4861 & 0.2586 & 6.23  & 25.55&  0.5813\\ 
& Ours(thr=0.35)   & 20.80&  0.3574& \underline{0.4708} & \underline{0.2374}  & \underline{ 5.35} & \underline{23.60} &\underline{0.5995}\\
& Ours(thr=0.40) &  20.56& 0.3460& \textbf{0.4693} & \textbf{0.2364}  & 5.37 & \textbf{22.30} & \textbf{0.6212}\\
\bottomrule
\end{tabular}
}

\end{table*}

\section{Experiments}

\begin{figure*}[t]
\centering
\includegraphics[width=2.1\columnwidth]{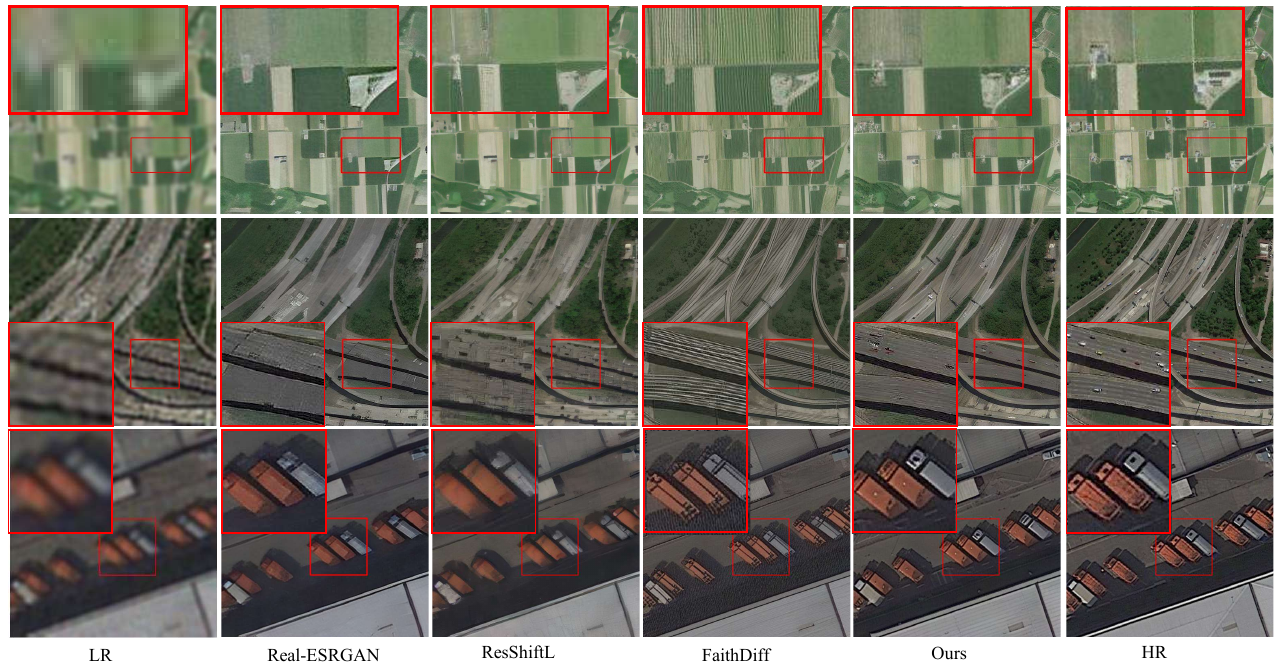} 
\caption{Image SR results (×4) on the synthetic scenario.}
\label{fig4}
\end{figure*}

\begin{figure*}[t]
\centering
\includegraphics[width=2.1\columnwidth]{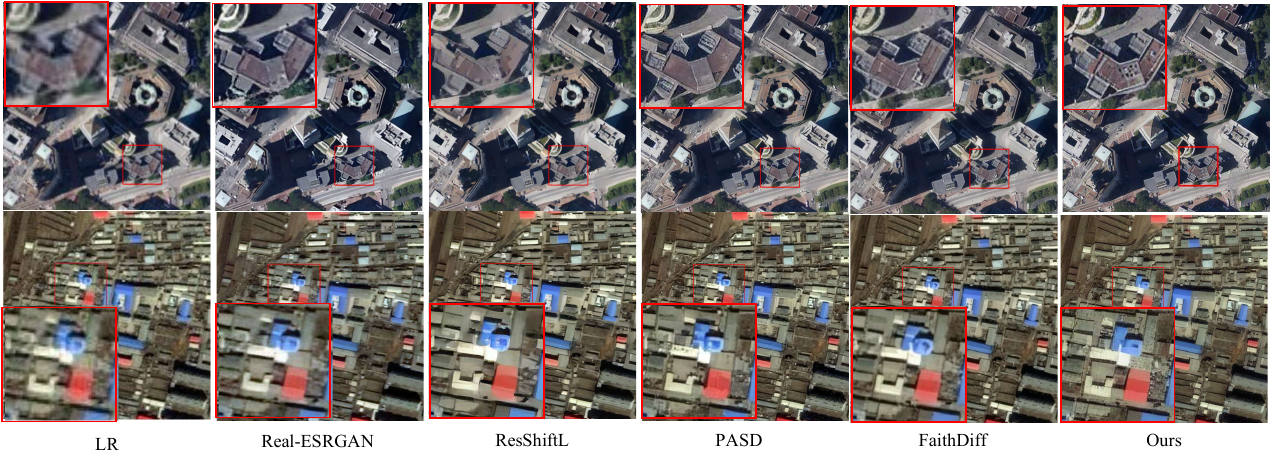} 
\caption{Image SR results (×4) on the Real-World scenario.}
\label{fig5}
\end{figure*}


\subsection{Experimental Settings}

\textbf{Datasets.} We construct our training dataset from the LoveDA-train~\cite{wang2021loveda}, DOTA-train~\cite{xia2018dota}, and a filtered subset of MillionAID~\cite{long2021creating}, resulting in a collection of approximately 300k remote sensing images. For LR images synthesis, we employ Real-ESRGAN degradation pipeline~\cite{wang2021real}, using the same configuration as FaithDiff~\cite{chen2025faithdiff}. Our synthetic evaluation dataset consists of the complete LoveDA~\cite{wang2021loveda} validation set, the full RSC11~\cite{zhao2016feature} dataset, 1,000 randomly selected images from AID~\cite{xia2017aid}, and 1,000 randomly cropped 512×512 patches from the DOTA~\cite{xia2018dota} test set. We degrade them using the same pipeline as the training data.
To evaluate the model’s performance on real-world remote sensing images, we select the full SIRI-WHU dataset~\cite{zhao2015dirichlet} (with a spatial resolution of 2m) as real-world test set, without applying any preprocessing or artificial degradation.

\textbf{Implementation Details.} We train TexADiff on our dataset for 30,000 steps with a batch size of 256. All baselines are retrained on our dataset. For all methods, we adopt the same degradation pipeline and keep configurations consistent across models. Since PASD~\cite{yang2024pixel} and FaithDiff~\cite{chen2025faithdiff} are also based on diffusion priors, we further align their training hyperparameters with those used in our approach. Detailed training settings are provided in the supplementary material. 

\textbf{Metric.} We adopt both reference metrics (PSNR, SSIM, LPIPS~\cite{zhang2018unreasonable}, and DISTS~\cite{ding2020image}) and no-reference metrics (NIQE~\cite{mittal2012no}, BRISQUE~\cite{mittal2012making}, and CLIP-IQA+~\cite{wang2023exploring}). PSNR and SSIM assess pixel-level similarity. LPIPS and DISTS are perceptual reference metrics. NIQE and BRISQUE are statistical no-reference metrics. CLIP-IQA+ serves as a perceptual no-reference metric leveraging vision-language priors. It is important to note that these no-reference metrics are designed for natural images and are often trained on natural image datasets. Due to the lack of widely established benchmarks for RSISR, we adopt them as auxiliary metrics. All metrics are reported on the synthetic test set, whereas only no-reference metrics are provided for the real-world test set due to the lack of ground-truth images.

\subsection{Comparisons with the State of the Art}


\begin{table}[t]
\centering
\small
\setlength{\tabcolsep}{1pt}
\caption{Quantitative comparison on the SIRI-WHU dataset. }
\label{tab:real}
\begin{tabular}{l|cccc} 
\toprule
Methods & NIQE$\downarrow$ & BRISQUE$\downarrow$ & CLIP-IQA+$\uparrow$ &User Study (\%)$\uparrow$ \\
\midrule
Real-ESRGAN        & 5.30              & 40.45              & 0.4099              & -- \\
ResShiftL          & 7.11              & 26.82              & 0.5314              & -- \\
PASD               & \textbf{4.50}     & \textbf{24.12}  & \textbf{0.5934}     & \underline{36.7} \\
FaithDiff          & 5.26              & 28.29              & 0.5034              & 16.9 \\
Ours(thr=0.35)     & 5.28              & 26.11              & 0.5424              & -- \\
Ours(thr=0.40)     & \underline{5.24}  & \underline{24.77}              & \underline{0.5628}     & \textbf{46.4} \\
\bottomrule
\end{tabular}


\end{table}

\begin{table}[htbp]
    \centering
    \caption{Comparison of complexity and inference efficiency.}
    \begin{tabular}{lccc}
        \toprule
        {\small\textbf{Method}} & {\small\textbf{Trainable}} & {\small\textbf{Total}} & {\small\textbf{Infer}} \\
        & {\small\textbf{Params (B)}} & {\small\textbf{Params (B)}} & {\small\textbf{Time (s)}} \\
        \midrule
        FaithDiff & 2.6 & 2.7 & 7.8 \\
        Ours      & 1.3 & 2.7 & 9.0 \\
        \bottomrule
    \end{tabular}
    \label{tab:model_effe}
\end{table}

\begin{table}[htbp]
\centering
\caption{Comparison of segmentation metrics on LoveDA.}
\begin{tabular}{lccc}
\toprule
Methods & Overall Accuracy$\uparrow$ & mIoU$\uparrow$ & mF1$\uparrow$ \\
\midrule
PASD  & 59.85 & \underline{41.33} & \underline{57.16} \\
FaithDiff & \underline{60.98} & 40.25 & 56.34 \\
Ours & \textbf{64.57} & \textbf{44.06} & \textbf{60.27} \\
\bottomrule
\end{tabular}
\label{tab:downstream}
\end{table}

We compare our method against state-of-the-art super-resolution approaches and commonly used baseline methods, including GAN-based models (Real-ESRGAN~\cite{wang2021real}), diffusion-based models (ResShiftL~\cite{yue2024efficient}, PASD~\cite{yang2024pixel} , FaithDiff~\cite{chen2025faithdiff}). 
Results obtained by retraining on the same dataset are reported in subsequent sections for a fair comparison. Additionally, results using their officially released checkpoints are provided in the supplementary material. 

\textbf{Evaluations on the Synthetic Datasets.}
We evaluate our method across multiple datasets, with quantitative results summarized in Table~\ref{tab:sy}. We report results using two different RTDM binarization thresholds during inference. Diffusion-based methods generally yield lower PSNR and SSIM scores, as they prioritize perceptual realism over strict pixel-level alignment with the ground truth. Notably, TexADiff consistently ranks among the top two in terms of reference-based perceptual metrics (LPIPS and DISTS) in most cases under both threshold settings.

Since most existing no-reference metrics are ill-suited for remote sensing images, no single method maintains a consistent lead across all datasets for these metrics. A detailed discussion regarding the limitations of no-reference metrics in evaluating remote sensing images quality is provided in the supplementary material.

Figure~\ref{fig4} presents qualitative comparisons in synthetic scenario. Our method demonstrates superior performance across diverse scenes and degradation conditions, producing semantically accurate structures without noticeable blurring. In contrast, GAN-based methods often produce textures that appear overly synthetic or lack structural consistency in certain regions. Other diffusion-based methods also yield suboptimal reconstruction results. ResShiftL generates structures that deviate from real-world geometries. FaithDiff tends to generate over-saturated textures in texture-sparse regions, such as agricultural fields.


\textbf{Evaluations on the Real-World Datasets.} The quantitative results are presented in Table~\ref{tab:real}. The qualitative comparisons are shown in Figure~\ref{fig5}; our approach produces sharper images with more realistic and faithful textures.  In addition, considering that existing no-reference metrics are derived from the natural image domain and may not be strictly applicable to remote sensing, as illustrated by concrete examples in the supplementary material, we further conducted a user study. Specifically, we invited 18 expert volunteers in remote sensing to select the “clearest and most faithful” result from 20 groups of randomly chosen real-world outputs produced by the most competitive methods. Our method achieved the best result in the user study.

\textbf{Efficiency and Complexity Analysis.} As shown in Table~\ref{tab:model_effe}, we compare our method with the primary competitor, FaithDiff~\cite{chen2025faithdiff} (both based on SDXL~\cite{podell2024sdxl}), in terms of parameters and inference time. Inference time is measured for processing a 1024×1024 image on a single RTX 3090 GPU, including the denoising process and additional components, but excluding the time for text generation in FaithDiff. Our approach outperforms the fully fine-tuned FaithDiff with fewer trainable parameters.

\subsection{Impact on Downstream Task}
To further evaluate the effectiveness of TexADiff on downstream task, we perform semantic segmentation on the LoveDA~\cite{wang2021loveda} dataset using DeepLabV3+~\cite{chen2018deeplabv3+} from the MMSegmentation~\cite{mmseg2020} with the official pretrained weights; the results are summarized in Table ~\ref{tab:downstream}, and our approach maintains the top position across all metrics.

\subsection{Ablation Study}
We conduct ablation study on the AID~\cite{xia2017aid} set to rigorously evaluate the effectiveness of our proposed method. To ensure a fair comparison, all experiments employ consistent training and inference settings, with modifications limited exclusively to the component being examined. Owing to computational resource limitations, for the ablation experiments that require training, we adopt a smaller batch size and fewer training iterations.

\textbf{The Effectiveness of Different Strategies.} We conduct an ablation study to evaluate the contribution of each component in our proposed TexADiff. As shown in Table~\ref{tab:diff_ablation}, "RTDM as cond" indicates the use of RTDM as a conditioning signal, and "TA-sampling" stands for Texture-aware Sampling. By incrementally incorporating each strategy, TexADiff achieves simultaneous improvements in both PSNR and LPIPS, indicating more accurate pixel-level reconstruction while also introducing richer perceptual details. This aligns well with the objective of our method: to generate richer and more faithful high-frequency details in texture-rich regions, while suppressing hallucinations in texture-sparse regions to achieve pixel-level fidelity. Notably, by employing the texture-aware sampling schedule, which reduces the number of sampling steps in texture-sparse regions, we observe even better performance in LPIPS and DISTS. This suggests that diffusion models benefit more from appropriate step allocation rather than simply increasing the number of denoising steps.

\begin{table}[t]
\centering
\caption{Ablation of different strategies.}
\setlength{\tabcolsep}{3pt}
\begin{tabular}{cccccc}
\toprule
\makecell{RTDM\\ as cond.} & TADL & TA-sampling. & PSNR↑ & LPIPS↓ & DISTS↓ \\
\midrule
--  & -- & -- & 21.86 & 0.4173 & 0.2042 \\
\checkmark  & -- & -- & 22.58 & 0.4023 & 0.2049   \\\
\checkmark & --   & \checkmark & 22.59 & 0.4001 & \underline{0.2039} \\
\checkmark & \checkmark   & -- & \textbf{22.78} & \underline{0.3887} & 0.2044 \\
\checkmark & \checkmark & \checkmark & \textbf{22.78} & \textbf{0.3883} & \textbf{0.2038} \\
\bottomrule
\end{tabular}

\label{tab:diff_ablation}
\end{table}

\textbf{Impact of Different RTDM Masks at Inference.} To assess how the choice of RTDM affects the final reconstruction quality, we perform an ablation study with five different masks:
(i) all‑ones mask, which treats the entire image as texture‑rich regions; (ii) all‑zeros mask, which treats the entire image as texture‑sparse regions; (iii) ground‑truth RTDM, obtained directly from each LR-HR pair by our texture density estimation method, and (iv) predicted RTDM, produced by our RTDM prediction network; and (v) an inverted RTDM, derived by flipping the predicted RTDM to swap texture-rich and texture-sparse regions. As shown in Table~\ref{tab:rtdm_ablation}, using the all-zeros mask yields the highest PSNR. This suggests that the zero values in the mask effectively guide the model to suppress the synthesis of fine details, thus favoring pixel-level fidelity. Directly employing the RTDM generated by our texture density estimation method achieves superior perceptual quality, highlighting the value of more accurate texture guidance. With continued improvements in RTDM prediction network design, further gains in reconstruction quality are expected. Notably, inverting the predicted RTDM—which misidentifies texture-rich and texture-sparse regions—leads to a significant performance drop, thereby validating the effectiveness of our current RTDM prediction network.

\begin{table}[t]
\centering
\caption{Ablation on the effect of different RTDM masks. }
\setlength{\tabcolsep}{2pt}
\begin{tabular}{lccc}
\toprule
\textbf{Mask Variant} & PSNR\,$\uparrow$ & LPIPS\,$\downarrow$ & DISTS\,$\downarrow$   \\
\midrule
All‑ones (full rich)           & 21.82 & 0.4098 & 0.2059  \\
All‑zeros (full sparse)        & \textbf{ 23.70 } & 0.4205 & 0.2142  \\
Inverted RTDM          & \underline{23.16} & 0.4305 & 0.2115  \\
Predicted RTDM (ours)   & 22.62 & \underline{0.3823} & \underline{0.1953}  \\
Ground‑truth RTDM (oracle) &22.84  & \textbf{0.3493} &  \textbf{0.1831}  \\
\bottomrule
\end{tabular}

\label{tab:rtdm_ablation}
\end{table}

\textbf{Selection of Unfrozen Parameters.} During fine-tuning, we unfreeze only a subset of the base model parameters. To verify the effectiveness of this design, we also evaluate several alternative configurations, such as unfreezing all downsampling blocks, the middle block, and the last upsampling block, or unfreezing only the attention layers. Table~\ref{tab:ablation_params} presents different results. Our unfreezing strategy achieves the best performance across all evaluation metrics.

\textbf{Impact of RTDM Binarization Threshold.} Table~\ref{thr} presents a comparison of different binarization schemes employed during both training and inference. First, the ablation of binarization leads to a substantial degradation across all evaluation metrics, underscoring its indispensable role in our framework. Furthermore, we investigate the impact of various threshold configurations by employing different binarization intervals during training and adjusting thresholds during inference. The results indicate that the current configuration serves as an empirically effective choice, achieving an optimal balance on this ablation subset.



\begin{table}[htbp]
    \centering
    \caption{Ablation of different unfreezing strategies.}
    \label{tab:ablation_params}
    \setlength{\tabcolsep}{3pt} 
    \begin{tabular}{lcccc}
        \toprule
        Model & Params (B) & PSNR$\uparrow$ & LPIPS$\downarrow$ & DISTS$\downarrow$ \\
        \midrule
        Cross-Attn & 0.10 & 20.79 & 0.6001 & 0.2952 \\
        Down+Mid & 1.25 & \underline{22.27} & \underline{0.4097} & \underline{0.2057} \\
        Ours & 1.33 & \textbf{22.78} & \textbf{0.3883} & \textbf{0.2038} \\
        \bottomrule
    \end{tabular}
\end{table}

\begin{table}[h]

\centering 
\caption{Ablation of different binarization thresholds.}
\setlength{\tabcolsep}{4pt} 
\begin{tabular}{l|l|ccc}
\toprule
Training  & Inference & PSNR$\uparrow$ & LPIPS$\downarrow$ & DISTS$\downarrow$ \\
\midrule

w/o {\scriptsize binarization}
& --      & 19.02 & 0.7269 & 0.3094\\
\midrule

\multirow{2}{*}{0.32--0.37} 
& 0.32      & 22.79 & 0.3953 & \underline{0.2025}\\
& 0.37      & 22.57 & 0.4006 & 0.2027\\
\midrule

\multirow{3}{*}{0.35--0.40} 
& 0.35 & \textbf{22.98}& 0.3903 & 0.2059\\
& 0.37 & \underline{22.88} & 0.3910& 0.2049 \\
& 0.40  & 22.78& \textbf{0.3883} & 0.2038\\
\midrule

\multirow{2}{*}{0.38--0.43} 
& 0.38      & 22.85&0.3937 &\textbf{0.2023} \\
& 0.43      & 22.66& 0.3959& 0.2031\\
\bottomrule
\end{tabular}
\label{thr}
\end{table}

%% file: sec/5_conclu.tex
\section{Conclusion}

In this work, we present TexADiff, which effectively addresses the imbalanced and spatially clustered texture distributions in remote sensing images. TexADiff reduces blurring and generates more faithful textures in texture-rich regions while suppressing hallucinated details in texture-sparse regions, achieving superior performance on both synthetic and real-world remote sensing datasets.
